\RequirePackage{fix-cm}
\documentclass[final, twocolumn,natbib]{svjour3}
\usepackage{appendix}
\usepackage{amsmath}
\usepackage{graphicx}
\usepackage{lineno}
\usepackage{array}
\usepackage{longtable}
\usepackage{url}
\usepackage{booktabs}
\usepackage[caption=false]{subfig}
\usepackage[misc,geometry]{ifsym} 
\usepackage{lipsum}
\usepackage{color}

\smartqed  
\usepackage{graphicx}
\begin{document}

\title{Multi-channel CNN to classify nepali covid-19 related tweets using hybrid features
}


\author{Chiranjibi Sitaula         \and
        Tej Bahadur Shahi 
}


\institute{C. Sitaula \at
            Department of Electrical and Computer Systems Engineering\\ Monash University\\
            Wellington Rd, Clayton VIC 3800, Australia\\
              \email{Chiranjibi.Sitaula@monash.edu}           
           \and
           TB. Shahi \at
             Central Department of Computer Science and Information Technology\\
              Tribhuvan University\\
               TU Rd, Kirtipur 44618, Kathmandu, Nepal \\
              \email{tejshahi@cdcsit.edu.np}\\
}

\date{Received: date / Accepted: date}

\maketitle

\begin{abstract}
Because of the current COVID-19 pandemic with its increasing fears among people, it has triggered several health complications such as depression and anxiety. Such complications have not only affected the developed countries but also developing countries such as Nepal. These complications can be understood from peoples' tweets/comments posted online after their proper analysis and sentiment classification.
Nevertheless, owing to the limited number of tokens/words in each tweet, it is always crucial to capture multiple information associated with them for their better understanding. 
In this study, we, first, represent each tweet by combining both syntactic and semantic information, called hybrid features. The syntactic information is generated from the bag of words method, whereas the semantic information is generated from the combination of the fastText-based (ft) and domain-specific (ds) methods. Second, we design a novel multi-channel convolutional neural network (MCNN), which ensembles the multiple CNNs, to capture multi-scale information for better classification. Last, we evaluate the efficacy of both the proposed feature extraction method and the MCNN model classifying tweets into three sentiment classes (positive, neutral and negative) on NepCOV19Tweets dataset, which is the only public COVID-19 tweets dataset in Nepali language. 
The evaluation results show that the proposed hybrid features outperform individual feature extraction methods with the highest classification accuracy of 69.7\% and the MCNN model outperforms the existing methods with the highest classification accuracy of 71.3\% during classification.
\end{abstract}

\section{Introduction}

%
{
COVID-19, which is also called SARS-COV2, is one of the deadliest viruses that have killed millions of people around the globe and still threatening to the humanity. The reason of death is not only attributed to its severe infections but also mental and psychological disorders triggered by its fear. Such disorders could be understood from the social media contents (e.g., tweets, posts, comments, etc.) provided by the users during this pandemic. The analysis of such contents could identify their mental status, thereby preventing from further health deterioration. 
}

%

\begin{figure}
    \centering
    \includegraphics[height=95mm, width=0.45\textwidth,keepaspectratio]{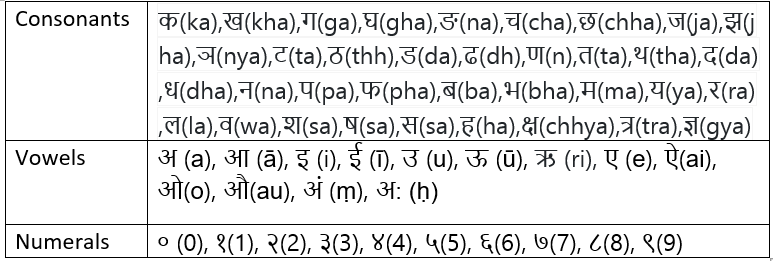}
    \caption{List of consonants, vowels and numerals of Nepali language \citep{sitaula2021deep}.}
    \label{fig:alphabets}
\end{figure}

There are several works conducted to analyze the COVID-19 tweets in non-Nepali \citep{boon2020public,de2021comparing,aljameel2021sentiment} and Nepali language \citep{sitaula2021deep,shahi2021natural}, which has a complex structure \citep{sitaula2013semantic,sitaula2012semantic}.
These research works in both Nepali and non-Nepali language utilize the basic feature extraction methods (e.g., term-frequency \& inverse document frequency, bag of words, etc.) only, which limit other discriminating information related to them. Particularly, tweets, which are shorter in length, need multiple information to distinguish them more accurately into positive, neutral and negative classes. Also, the feature extraction methods and classification models employed in non-Nepali, which are mostly conducted in high-source language, might not be applicable to the low-resource languages such as Nepali language. Nepali language is based on Devanagari script, which contains 36 consonants. Among them, 33 are distinct consonants, whereas 3 are combined consonants), 13 vowels, and 10 numerals (see the details in Fig. \ref{fig:alphabets}). 

%
{
To address the aforementioned limitations, we perform the following tasks. We propose to use both syntactic (bag of words) and semantic information (domain-specific \citep{sitaula2021deep} and fastText-based \citep{bojanowski2017enriching}). The syntactic information help capture the token occurrence patterns within input tweet and tweets among other classes. Furthermore, the semantic information help capture contextual information. The combination of both syntactic and semantic information helps better understand the tweet in multiple ways and work complementary to each other.
Furthermore, the discriminability of each tweet could further be improved by capturing the multi-scale features, which can be achieved by a multi-scale convolution neural network(MCNN). For this reason, we propose the novel MCNN model for the classification purpose.
}
%

{
Herein, our paper possesses the following {\bf contributions}:
\begin{itemize}
    \item[(i)] propose a novel multi-channel convolutional neural network, called the MCNN model, to train in an end-to-end manner for the classification of the Nepali COVID-19 related tweets into three different sentiment classes (positive, neutral, and negative);
    \item[(ii)] propose a novel hybrid feature extraction method combining both syntactic (bag of words) and semantic (domain-specific and fastText-based) in the Nepali language context; 
    and
    \item[(iii)] demonstrate the superiority in terms of well-established metrics of our proposed feature extraction method and the MCNN model on NepCov19Tweets dataset. 
\end{itemize}
}

{The overall paper is organized as follows. Section \ref{related_works} mentions the related works of our study, including both Nepali and non-Nepali language processing. Furthermore, Section \ref{materials_methods} presents the required resources and the proposed methods, which is followed by the experimental results and their discussion in Section \ref{results_discussion}. Finally, the paper is concluded in Section \ref{conclusion_futureworks}.}

\section{Related works}
\label{related_works}
We study the COVID-19 related tweets classification for sentiment analysis in terms of two different aspects: non-Nepali and Nepali.
Although there are some more works carried out in non-Nepali language, very few works have been carried out in Nepali language.

Under non-Nepali language, COVID-19 tweets sentiment classification tasks were carried out in different languages (e.g., Arabic, English, etc.).
For example, \cite{boon2020public} released the public dataset of COVID-19 related tweets for the sentiment classification in English.
{At first, \cite{chandrasekaran2020topics} and  \cite{xue2020public} employed the topic modeling approach over the COVID-19 tweets to identify the tweets. However, \cite{rustam2021performance} used the traditional machine learning classifiers for the classification of tweets into three different classes, which achieves a classification accuracy of 93.00\% with Extreme Tree classifier (ETC). At the same time, researchers in \cite{kaur2021proposed} employed the hybrid heterogeneous support vector machine (HH-SVM) to classify the COVID-19 tweets, which outperforms the recurrent neural network classifier (RNN).
Considering the efficacy of different machine learning classifiers, \cite{naseem2021covidsenti} used different classifiers and found that fine-tuned BERT model imparts a classification accuracy of 92.90\% during the classification of COVID-19 related tweets. Similarly, \cite{basiri2021novel} developed the ensemble deep learning model to classify the COVID-19 related tweets into different sentiment classes, which produce a classification accuracy of 85.80\%.}
Furthermore, \cite{de2021comparing} utilized the sentiment classification on Brazilian tweets related to COVID-19, whereas \cite{aljameel2021sentiment} conducted the sentiment analysis on COVID-19 tweets collected in Arabic language, which provides a classification accuracy of 85\% with the support vector machine classifier.
For the classification of tweets into different sentiment classes (e.g., positive, neutral and negative), they used well-established popular classifiers such as eXtreme gradient boosting (XGBoost), Long short-term memory (LSTM) and Extra tree classier. For the feature extraction, they used popular feature extraction methods such as bag of words and word embedding.

Under Nepali language, there is only one work carried out recently to classify and represent the COVID-19 tweets into three different sentiment classes (positive, neutral and negative). For instance, \cite{sitaula2021deep} developed a novel approach to classify the Nepali COVID-19 related tweets into three sentiment classes (negative, positive and neutral). They used domain-specific (ds), domain-agnostic (da), and fastText (ft) based information for the representation of each tweet and trained individual CNNs for each of them. Finally, the ensemble model was designed to exploit all three kinds of information for the classification, which provides the classification accuracy of 68.7\%.
In addition, there are some closely related works carried out in Nepali language. For example, \cite{shahi2018nepali} and  \cite{basnet2018improving} represented the Nepali texts using bag of words and classified using the support vector machine classifier (classification accuracy: 74.65\%) and LSTM (classification accuracy: 84.63\%), respectively. Similarly, \cite{subba2019nepali} employed bag of words approach to represent the Nepali text and classify them using deep neural network. More recently, \cite{sitaula2021vector} designed a supervised codebook approach to represent the Nepali texts and classified them using support vector machine classifier, which provides an accuracy of 89.58\%. It is noticed that bag of words (BoW) and traditional machine learning (ML) classifiers have been preferred for the  Nepali text representation and classification, respectively. However, there is still lacking in the use and adoption of contextual information to enrich the representational ability of Nepali texts to achieve an excellent classification performance.

\begin{figure*}[!htbp]
    \centering
    \includegraphics[height=95mm, width=0.95\textwidth,keepaspectratio]{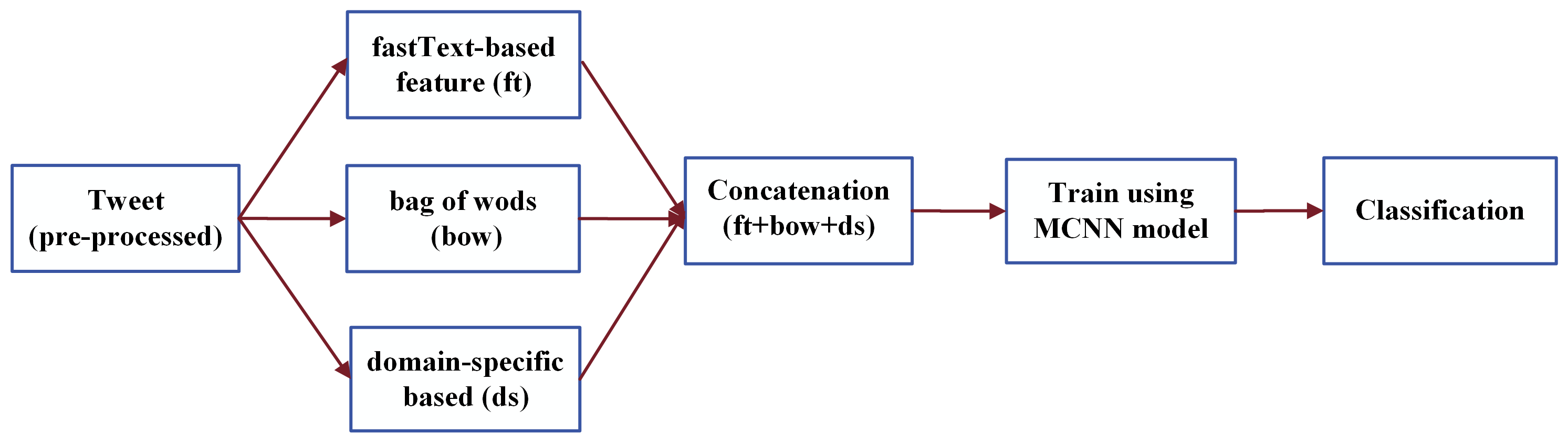}
    \caption{Overall pipeline of our proposed approach. Pre-processed tweets are used to extract the features using fastText-based (ft), bag of words (bow), and domain-specific (ds) method. Then after, we concatenate them, called hybrid features, to train our proposed MCNN model for the classification.}
    \label{fig:pipeline}
\end{figure*}

\section{Materials and methods}
\label{materials_methods}

\subsection{Dataset}

{We use the dataset, called NepCov19Tweets \citep{sitaula2021deep}, in this study, which has three challenging classes--positive, neutral, and negative-- having a higher inter-class similarity and intra-class dissimilarity problems.
This dataset is the only publicly available dataset to perform the sentiment analysis of Nepali COVID-19 related tweets, which has 33,247 total tweets (positive class: 14,957 samples;  neutral class: 4,744 samples; and negative class: 13,546 samples).
}

\subsection{Evaluation metrics}
{We evaluate our method using widely-used performance metrics such as Precision (Eq. \eqref{eq:precision}), Recall (Eq. \eqref{eq:recall}), F1-score (Eq. \eqref{eq:f-score}), and Accuracy (Eq. \eqref{eq:acc}).}

\begin{equation}
      P= \frac{TP}{TP+FP},
    \label{eq:precision}
    \end{equation}

\begin{equation}
      R= \frac{TP}{TP+FN},
    \label{eq:recall}
    \end{equation}
    
    \begin{equation}
      F= 2 \times \frac{P \times R}{P+R},
    \label{eq:f-score}
    \end{equation}
   
    \begin{equation}
    A = \frac{TP+TN}{TP+TN+FP+FN},
    \label{eq:acc}
    \end{equation}
where $TP$, $TN$, $FP$, and $FN$ represent true positive, true negative, false positive, and false negative, respectively. Similarly, $P$, $R$, $F$, and $A$ represent Precision, Recall, F1-score, and Accuracy, respectively.

\subsection{Machine learning algorithms}
{In this study, we evaluate and compare eight popular machine learning baseline algorithms with our proposed multi-channel convolutional neural network classifier (MCNN). They are Support Vector Machine (SVM)  (both linear and radial basis function) \citep{cristianini2000SVM}, Random Forest (RF) \citep{randomClassifier}, eXtreme Gradient Boosting (XGBoost) \citep{xgboost}, Naive Bayes (NB) \citep{pedregosa2011scikit}, Artificial Neural Network (ANN) \citep{dimililer2013backpropagation}, Logistic Regression (LR) \citep{fan2008liblinear}, and K-Nearest Neighbors (K-NN) \citep{mucherino2009k}.
The selection of these algorithms ranges from the traditional machine learning algorithms (e.g., SVM) to the latest algorithms (e.g., ANN, XGBoost, etc). }

\subsection{Proposed approach}
In our proposed method, we follow four steps: \ref{step_1} Pre-processing; \ref{step_2} Feature extraction; \ref{step_3} Four CNNs design for fine-tuning; \ref{step_4} Multi-channel CNN design for the classification.
The high-level diagram of our overall pipeline is presented in Fig. \ref{fig:pipeline}. 

\subsubsection{Pre-processing}
\label{step_1}
We use three different steps to pre-process each tweet. First, we remove the stop words, which are unimportant to capture the contextual information. Second, we remove the alphanumeric characters in the tweets.  Third, we perform the stemming operation to extract the root word, which eliminates the redundant words.

\subsubsection{Feature extraction}
\label{step_2}
After the pre-processing operations, we extract four different features in this study. First, we extract fastText-based features (ft), which is 300-D size and domain-specific features (ds) \cite{sitaula2021deep}, which is 3-D size.
The 'ft' based features are based on a pre-trained model, which has been pre-trained with textual data from diverse disciplines.
Similarly, the 'ds' based features are based on the probability of each token under three different sentiment classes (negative, neutral and positive). Here, based on the probability of each token among three different categories, we identify the importance of each token.
In this study, we combine both 'ft' and 'ds' to exploit the semantic information. Furthermore, for the syntactic information, we exploit the popular algorithm, called bag of words (bow) approach, which employs the frequency and inverse document frequency technique \citep{shahi2018nepaliSMS}.
Finally, we combine all three different kinds of features to establish the hybrid features representing each tweet, which are used to train our CNNs in this study (refer to Eqs. \eqref{ft:feature},\eqref{bow:feature},\eqref{ds:feature}, and \eqref{hybrid} denote 'ft', 'bow', 'ds', and 'hybrid' features in this study, respectively).

\begin{equation}
 ft=fastText(d),
    \label{ft:feature}
\end{equation}

\begin{equation}
bow=BagOfWords(d),
    \label{bow:feature}
\end{equation}

\begin{equation}
ds=DomainSpecific(d),
    \label{ds:feature}
\end{equation}

\begin{equation}
h=[ft,bow,ds],
    \label{hybrid}
\end{equation}
where $d$ and $h$ denote the input tweet and hybrid features (proposed), respectively. Also, fastText(.), BagOfWords(.), and DomainSpectific(.) denote the fastText-based ('ft'), bag of words-based method ('bow'), and domain-specific-based feature extraction ('ds') methods to represent each tweet in this study.
To achieve the 'ft' features in this study, we extract the 300-D features for each token and average them to obtain the final representation for each tweet ($d$). Similarly, we consider top-100 tokens to obtain the 100-D bag of words feature size, whereas for the 'ds' feature, we extract 3-D feature size for each token and average them representing each tweet ($d$). Finally, we concatenate all three of them to attain the hybrid features ($h$) of 403-D to train our CNNs.

\subsubsection{Four CNNs design and fine-tuning}
\label{step_3}
We design four different CNNs ($C_{1}$, $C_{2}$, $C_{3}$, and $C_{4}$ with the kernel of 1, 2, 3, and 4, respectively) to capture the multi-scale information. We fine-tune each CNN before establishing our proposed multi-channel CNN in this study. The detailed architecture of the proposed CNNs are shown in Table \ref{tab:architecture}.

\begin{table*}[!htbp]
    \centering
    \scriptsize
      \caption{Architecture of all four proposed CNNs representing four different channels in this study, where f and k represent the number of filters and kernel size, respectively.}
    \begin{tabular}{p{3cm}|c|c|c|c|c|c|c|c}
    \toprule
      &\multicolumn{2}{c|}{$C_1$} 
      &\multicolumn{2}{c|}{$C_2$} 
      &\multicolumn{2}{c|}{$C_3$}
      &\multicolumn{2}{c}{$C_4$}\\
      \cline{2-9}
         Layer&(f, k) &Output shape&(f, k) &Output shape &(f, k) &Output shape &(f, k) &Output shape \\
         \midrule
         Input  & - &(403,1)& - &(403,1) &-&(403,1)&- &(403,1)  \\     
         Conv1D+Relu&(32,1)&(403,32)&(32,2) &(402,32) &(32,3) &(401,32)&(32,4) &(400,32) \\
         Conv1D+Relu+Dropout (0.4)&(16,1) &(403,16) &(16,2) &(402,16) &(16,3) &(401,16)&(16,4) &(400,16)  \\
         MaxPooling1D (pool size=1)&-&(403,16) &- &(402,16) &- &(401,16)&-&(400,16)\\
         Flatten+Dropout (0.5)&- &6448 &- &6432 &- &6416&- &6400  \\
         Dense+Dropout (0.5)&- &128 &- &128 &- &128&- &128  \\
         Dense+Dropout (0.5)&- &64 &- &64 &- &64&- &64  \\
         Softmax&- &3 &- &3 &- &3&- &3  \\
          \bottomrule
    \end{tabular}
    \label{tab:architecture}
\end{table*}

 \begin{figure*}[!htbp]
    \centering
    \includegraphics[height=95mm, width=0.90\textwidth,keepaspectratio]{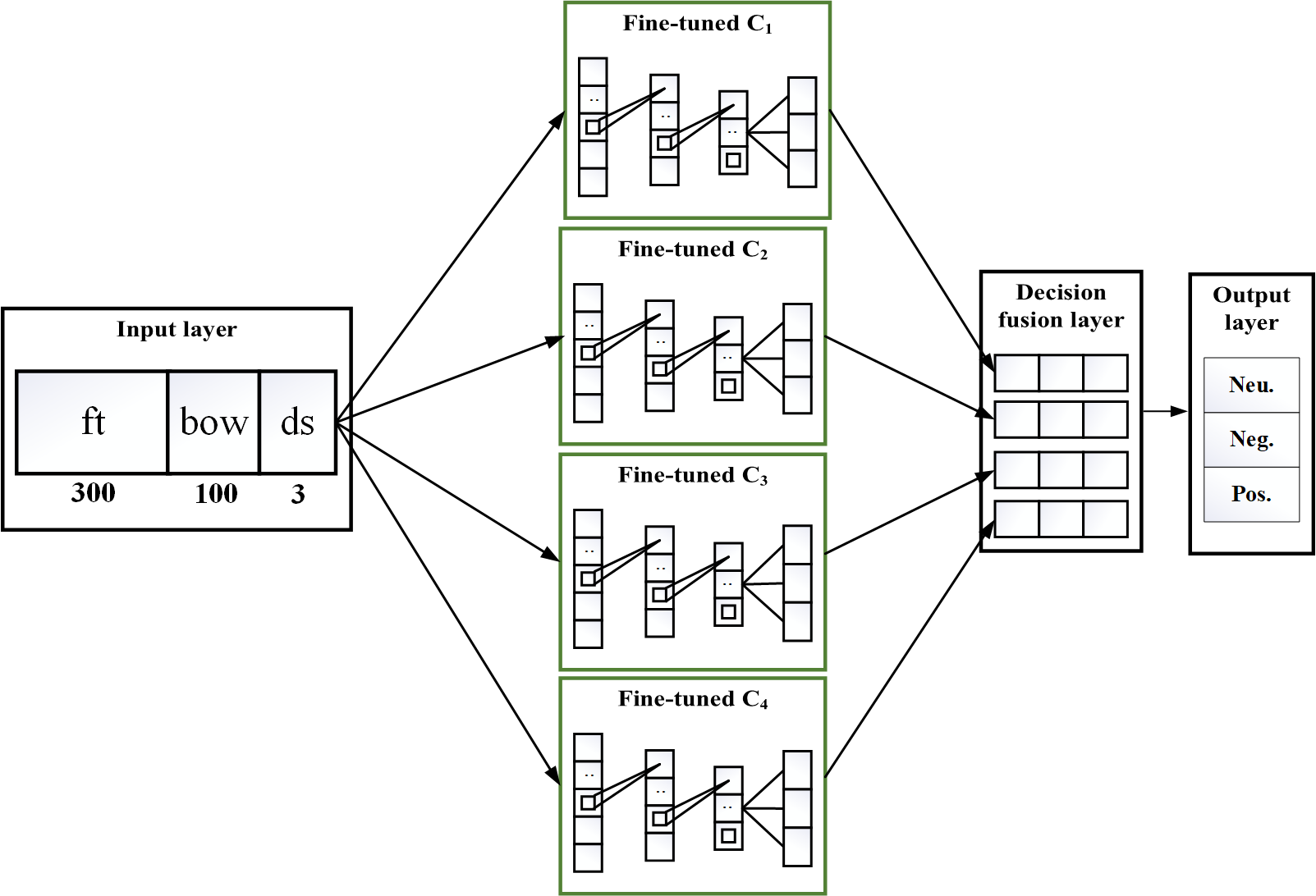}
    \caption{Proposed MCNN model with three classes (Neg: negative, Neu: neutral, and Pos: positive). Here, $C_{1}$, $C_{2}$, $C_{3}$, and $C_{4}$ denote CNN with kernel size of 1, 2, 3, and 4, respectively.}
    \label{fig:mcnn}
\end{figure*}

\subsubsection{Multi-channel CNN design and classification}
\label{step_4}
After fine-tuning each CNN model, we aggregate them using decision layer and train it in an end-to-end fashion for the classification. The aggregation in this approach not only helps reduce the number of trainable parameters but also improve the overall performance. 
The high-level diagram of our proposed MCNN is shown in Fig. \ref{fig:mcnn}.
The mathematical representation of output of each channel CNN and their fusion is shown in Eqs. \eqref{eq:c1}, \eqref{eq:c2}, \eqref{eq:c3}, \eqref{eq:c4},  and \eqref{eq:decision_fusion} for each CNN $C_{1}$, $C_{2}$, $C_{3}$, and $C_{4}$ and decision fusion, respectively.

\begin{equation}
C_1(a_i) = \frac{e^{a_{i}}}{\sum_{j=1}^K e^{a_{j}}} 
    \label{eq:c1}
\end{equation}

\begin{equation}
C_2(b_i) = \frac{e^{b_{i}}}{\sum_{j=1}^K e^{b_{j}}},
    \label{eq:c2}
\end{equation}

\begin{equation}
C_3(c_i) = \frac{e^{c_{i}}}{\sum_{j=1}^K e^{c_{j}}},
    \label{eq:c3}
\end{equation}

\begin{equation}
C_4(d_i) = \frac{e^{d_{i}}}{\sum_{j=1}^K e^{d_{j}}},
    \label{eq:c4}
\end{equation}

\begin{equation}
E(z_i)=Fusion\{C_1(a_i),C_2(b_i),C_3(c_i),C_4(d_i)\},
    \label{eq:decision_fusion}
\end{equation}

where $i\in \{1,2,3\}$. Here, $a_i$, $b_i$,$c_i$, and $d_i$ is the multinomial output value for $C_{1}$, $C_{2}$, $C_{3}$, and $C_{4}$ for index $i$, respectively.
For the decision fusion layer ($E(Z_i)$), we employ an average voting approach, which has produced prominent accuracy than other methods in this study (see details in Section \ref{sota_comparison}).

\begin{table*}[tb]
    \centering
       \caption{Comparison of different feature extraction methods in terms of averaged classification performance (\%).  Note that P, R, F, and A denote overall Precision, Recall, F1-score, and Accuracy for four types of methods (ft: fastText; ds: domain-specific; bow: bag of words; proposed (hybrid): ft+ds+bow), respectively. The hyper-parameters of traditional machine learning algorithms are as follows:  SVM+Linear (c: 1, Gamma: 0.1), SVM+RBF (c: 100, Gamma: 0.1), XGBoost (learning-rate: 0.1, max-depth: 7, n-estimators: 150), ANN (Hidden-layer-size: 20, learning-rate-init: 0.01, max-iter: 1000), RF (min-sample-leaf: 3, min-sample-split: 6, n-estimators: 200), LR (C: 10, solver: lbfgs, max-iter: 1000), K-NN (leaf-size: 35, n-neighbor: 120, p: 1). Boldface denotes the highest performance under the precision, recall, f1-score, and accuracy.
    }
    \scriptsize
    \begin{tabular}{p{1.7cm}|c|c|c|c|c|c|c|c|c|c|c|c|c|c|c|c}
    \toprule
    &\multicolumn{4}{c}{\bf ft}&\multicolumn{4}{|c}{\bf ds}&\multicolumn{4}{|c}{\bf bow } &\multicolumn{4}{|c}{\bf proposed}\\
    \cline{2-17}
     Algorithms & P & R&F&A &P & R & F& A& P&R&F&A&P&R&F&A\\
     \midrule
        NB&51.3&49.9&49.3&60.9&43.8&38.0&33.9&49.9&51.3&49.4&48.7&54.5&\bf 53.7&\bf 51.2&\bf 51.2&\bf 61.7\\
        ANN&57.0&56.9&56.9&63.7&\bf 58.9&49.5&48.1&62.0&52.1&50.7&50.8&59.5&58.1&\bf 57.9&\bf 57.9&\bf 64.8\\
         K-NN&\bf 64.0&50.6&47.4&65.1&57.9&48.4&46.5&61.5&58.9&47.3&45.2&60.3&63.5&\bf 51.0&\bf 47.8&\bf 65.6\\
         SVM+Linear&50.8&50.3 &50.3 & 59.4& \bf 59.3&47.1&43.4&\bf 61.4&49.8&47.4&46.1&59.4&52.4&\bf 51.8&\bf 51.9&60.4\\
        LR&60.7&55.0&54.6&67.3&61.5&48.5&46.4&61.8&59.1&49.7&47.8&63.0&\bf 62.3&\bf 57.7&\bf 58.2&\bf 68.6\\
        RF &71.5&53.6&52.0&67.4&55.5&50.7&50.5&61.8&59.1&50.7&49.5&63.5&\bf 71.7&\bf 53.8&\bf 52.3&\bf 67.7\\
         SVM+RBF &\bf 72.7 &36.2&26.7& 47.9&60.8&\bf 48.4&\bf 46.4&\bf 61.7&51.8&41.7&39.0&51.3&72.2&36.1&26.5&47.7\\
         XGBoost&67.9&57.2&57.3&69.4&58.4&49.8&48.7&62.2&60.0&50.6&49.3&63.5&\bf 68.0&\bf 57.6&\bf 57.8&\bf 69.7\\
        \midrule
       Average&61.9&51.2&49.3&62.6&57.0&47.5&45.4&60.2&55.2&48.4&47.0&59.3&\bf 62.7&\bf 52.1&\bf 50.4&\bf 63.2\\
       
         \bottomrule
    \end{tabular}
    \label{tab:comparison_sota}
\end{table*}

\subsection{Implementation}
To implement the traditional machine learning classifiers and our proposed model, we employ Sklearn \citep{scikit-learn},  and Keras \citep{chollet2015keras}, respectively implemented in Python \citep{python}. 
For the experimental evaluations, we design 10 different train/test splits, each with the ratio of 70/30 per category and report the averaged performance.
In the meantime, the optimal hyper-parameters of the traditional machine learning classifiers are taken from the recently published article using the NepCOV19Tweets dataset \citep{sitaula2021deep}.


\begin{table}[tb]
    \centering
    \scriptsize
        \caption{Comparison of our method against the state-of-the-art methods using averaged classification accuracy (\%) on NepCov19Tweets dataset. Note that 'avg', 'sum', and 'max' denote the mean pooling-based, sum-based, and maximum voting-based, respectively used in our proposed MCNN model.}
    \begin{tabular}{p{4cm}|c|c}
    \toprule
     Methods& Feature size & Accuracy\\
     \midrule
         \cite{shahi2018nepali}, 2018&100-D&62.1\\
         \cite{basnet2018improving}, 2018&300-D&62.9\\
         \cite{sitaula2021vector}, 2021 &17-D&59.8\\
        {\cite{sitaula2021deep}(ds), 2021}&3-D&61.5\\
        {\cite{sitaula2021deep} (da), 2021}&17-D&59.5\\
        {\cite{sitaula2021deep} (ft), 2021}&300-D&68.1\\
        {\cite{sitaula2021deep} (ds+da+ft), 2021}&320-D&68.7\\
        \midrule
        {\bf Ours (avg) }&403-D &\bf 71.3 \\
        {\bf Ours (sum) }&403-D &\bf 71.2 \\
        {\bf Ours (max) }&403-D &\bf 70.7 \\
        
         \bottomrule
    \end{tabular}
    \label{tab:comparison_sota_}
\end{table}

\begin{table*}[tb]
    \centering
    \scriptsize
    \caption{Class-wise study of our proposed MCNN method and its components using the averaged classification performance (\%). Note that P, R, F denote precision, recall, and f1-score for three classes (positive, negative and neutral), respectively. Note that the hyper-parameters used in our models are as follows: learning-rate: 1e-05 , batch-size: 32, epochs: 50 and optimizer: RMSProp. Boldface denotes the significant performance.}
    \begin{tabular}{c|c|c|c|c|c|c|c|c|c|c|c|c|c}
    \toprule
      & \multicolumn{3}{c|}{Negative}& \multicolumn{3}{c|}{Neutral} &\multicolumn{3}{c}{Positive}& \multicolumn{4}{|c}{Overall Average}\\
      \cline{2-14}
         CNN&P  &R &F &P &R &F &P &R&F&P&R&F&A \\
         \midrule
         $C_{1}$  &72.1 & 76.5 &74.1 &58.9 &20.4 &27.0 &68.8&81.9&74.8&66.6&57.2&56.9&69.9\\ 
          $C_{2}$&73.4 &75.2 &74.3 &59.9 &15.8 &25.0 &68.8 &\bf 83.5&75.4&67.3&58.1&58.2&70.3 \\
          $C_{3}$&73.2  &75.7 &74.5 &59.9 &16.4 &25.6 &69.0 &83.1&75.4&67.4&58.3&58.4&70.5 \\
          $C_{4}$&73.3  &75.1 &74.2 &\bf 69.4 &16.1 &26.4 &68.8 & 83.3&75.3&\bf 67.8&58.3&58.6&70.4 \\
          \midrule
          MCNN&\bf 73.5&\bf 76.6 &\bf 75.0 &56.6 &\bf 24.4 &\bf 33.9&\bf 71.4&81.4&\bf 76.0 &67.1 & \bf 60.8&\bf 61.6&\bf 71.3\\          \bottomrule
    \end{tabular}
    \label{tab:classwise_study}
\end{table*}

\section{Results and discussion}
\label{results_discussion}

\subsection{Comparison of proposed feature extraction methods using well-established ML methods}

We compare our proposed hybrid features using averaged classification performance (\%) with each individual feature ('ft', 'ds', and 'bow') against eight (including Linear and RBF SVM algorithms) well-established machine learning (ML) classifiers, which are presented in Table \ref{tab:comparison_sota}. 

While looking at Table \ref{tab:comparison_sota}, the proposed features (hybrid) impart the most promising performances compared to 'ft', 'ds', and 'bow' based features in most cases.  In terms of averaged classification accuracy, the proposed features outperform the other three individual features under six different ML classifiers. As an example, our method yields 69.7\% ( 0.3\% higher than second-best method, 'ft'), 67.7\% (0.3\% higher than second-best method, 'ft'), 68.6\% (1.3\% higher than second-best, 'ft'), and 65.6\% (0.5\% higher than second-best method, 'ft'), 64.8\% (1.1\% higher than second-best method, 'ft'), and 61.7\% (0.8\% higher than second-best method,'ft') from XGBoost, RF, LR, and, K-NN, ANN, NB, respectively. In terms of averaged precision, our proposed features outperform three remaining features against four different classifiers. For instance, it imparts 68.0\% (0.1\% higher than second-best method, 'ft'), 71.7\% (0.2\% higher than second-best method, 'ft'), 62.3\% (0.8\% higher than second-best method, 'ds'), and 53.7\% (2.4\% higher than second-best method, 'bow') from XGBoost, RF, LR, and NB, respectively.
Under the averaged recall, our proposed features outperform the remaining features for seven ML classifiers. 
For instance, it provides 57.6\% (0.4\% higher from second-best method, 'ft') from XGBoost, 53.8\% (0.2\% higher than second-best method, 'ft') from RF, 57.7\% (2.7\% higher than second-best method, 'ft') from LR, 51.8\% (1.5\% higher than second-best method, 'ft') from SVM+Linear, 51.0\% (0.4\% higher than second-best method, 'ft') from K-NN, 57.9\% (1.0\% higher than second-best method, 'ft') from ANN, and 51.2\% (1.3\% higher than second-best method, 'ft') from NB classifiers.
Finally, under the averaged F1-score, our proposed features surpass the remaining features while using seven ML classifiers. For example, it provides 57.8\% (0.5\% higher from second-best, 'ft') from XGBoost, 52.3\% (0.3\% higher than second-best method, 'ft') from RF, 58.2\% (3.6\% higher than second-best method, 'ft') from LR, 51.9\% (1.6\% higher than second-best method, 'ft') from SVM+Linear, 47.8\% (0.4\% higher than second-best method, 'ft') from K-NN, 57.9\% (1.0\% higher than second-best method, 'ft') from ANN, and 51.2\% (1.9\% higher than second-best method, 'ft') from NB classifiers.

{While analyzing the performance of the proposed feature extraction method using the averaged performance across all 8 classifiers, we notice that our method outperforms other individuals. Specifically, under precision, our method provides 62.7\%, which is 0.8\% higher that second-best method ('ft') and 5.7\% higher than least-performing method ('bow'). Similarly, our method imparts 52.1\% recall, which is 0.9\% higher than second-best method ('ft') and 4.6\% higher than least-performing method ('ds'). Furthermore, under average f1-score and accuracy, our method imparts 50.4\% (1.1\% higher than second-best method, 'ft' and 2.9\% higher than least-performing method, 'ds') and 63.2\% (0.6\% higher than second-best method, 'ft' and 3.9\% higher than least-performing method, 'bow'), respectively.
}

From the discussions above, we suggest that the proposed method surpasses the remaining three individual features with stable performance in terms of four different evaluation measures. We also note that the main driving force to improve the performance in our proposed hybrid features is fastText-based features ('ft') in our study. The reason for this is that it has been trained with a massive corpus from the diverse domains.
In addition, we believe that our features are suitable for XGBoost classifier to attain the best performance compared to other remaining ML classifiers. 

\subsection{Comparison of proposed MCNN model with state-of-the-art methods}
\label{sota_comparison}

Here, we compare our proposed MCNN model with the existing state-of-the-art methods in terms of the averaged classification accuracy, which are presented in Table \ref{tab:comparison_sota_}. 
Given that our MCNN model aggregates four different CNNs using three different methods--avg, sum, and max, we compare the classification accuracies of each of them with the state-of-the-art methods proposed in Nepali text classification. 

While looking at Table \ref{tab:comparison_sota_}, we observe that our proposed MCNN model with the 'avg' decision fusion imparts an accuracy of 71.3\%, which is 11.8\% higher than least-performing method (da) and 2.6\% higher than second-best method (ds+da+ft). Furthermore,  our proposed model with 'sum' and 'max' provide an accuracy of 71.2\% (11.7\% higher than least-performing method (da) and 2.5\% higher than second-best method (ds+da+ft) and 70.7\%  (11.2\% higher than least-performing method (da) and 2.0\% higher than second-best method (ds+da+ft)) respectively.

By and large, we believe that our proposed model has the ability to produce highly discriminant features during the classification of tweets into positive, neutral and negative classes. We find that 'avg' based decision fusion outperforms 'sum' and 'max' based decision fusion. The reason of stable performance from 'avg' is that it helps preserve both higher and lower probability values during decision fusion.

\subsection{Ablative study of MCNN model}

To show the efficacy of our method towards class-wise performance and contribution of each channel, we evaluate three different performance measures for three different classes for each channel and MCNN, which are presented in Table \ref{tab:classwise_study}.

While looking at Table \ref{tab:classwise_study}, we observe that our model imparts the highest precision (73.5\%, and 71.4\% for negative and positive, respectively), recall (76.6\%, and 24.4\% for negative and neutral, respectively), and f1-score (75.0\%, 33.9\%, and 76.0\% for negative, neutral and positive, respectively).
The performance improvement in our method is attributed to each channel that collected highly discriminating information. As an example, $C_1$ imparts precision, recall, and f1-score of 72.1\%, 76.5\%, and 74.1\% for negative class; 58.9\%, 20.4\%, and 27.0\% for neutral class; and 68.8\%, 81.9\%, and 74.8\% for positive class, respectively.
Furthermore, $C_2$ imparts precision, recall, and f1-score of 73.4\%, 75.2\%, and 74.3\% for negative class; 59.9\%, 15.8\%, and 25.0\% for neutral class; and 68.8\%, 83.5\%, and 75.4\% for positive class, respectively.
Furthermore, $C_3$ imparts precision, recall, and f1-score of 73.2\%, 75.7\%, and 74.5\% for negative class; 59.9\%, 16.4\%, and 25.6\% for neutral class; and 69.0\%, 83.1\%, and 75.4\% for positive class, respectively.
Finally, we achieve the precision, recall, and f1-score of 73.3\%, 75.1\%, and 74.2\% for negative class; 69.4\%, 16.1\%, and 26.4\% for neutral class; and 68.8\%, 83.3\% and 75.3\% for positive class in $C_4$, respectively. 
This result infers that our MCNN model outperforms its individual channel CNNs.

We also analyze the efficacy of each component used in our model in terms of overall averaged performance metrics (\%) including classification accuracy, which also reveal that our model outperforms individual channel CNNs in terms of all metrics. For example, our model provides precision, recall, f1-score, and accuracy of 67.1\%, 60.8\%, 61.6\%, and 71.3\%, respectively, which are the highest measures except precision among all individual channel CNNs. These encouraging results support that our MCNN model is able to preserve the discriminating information from individual CNN, thereby improving the overall classification performance significantly.

\begin{figure}
    \centering
    \includegraphics[height=90mm, width=0.46\textwidth,keepaspectratio]{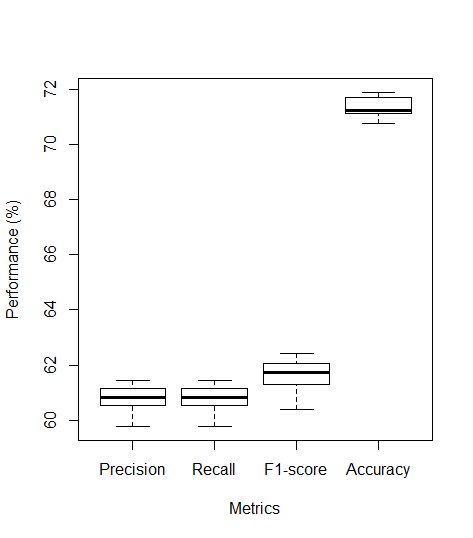}
    \caption{Box plots over the results presented using evaluation measures (precision, recall, f1-score and accuracy) for 10-fold splits. }
    \label{fig:boxplots}
\end{figure}

{\subsection{Statistical analysis}
Here, we perform the statistical analysis of the evaluation results achieved from our proposed method across 10-fold splits. For this, we employ the confidence interval, p-value, and box plot visualization of the results presented using evaluation measures (precision, recall, f1-score, and accuracy). The box plot visualization is shown in Fig. \ref{fig:boxplots}. 
Furthermore, our method imparts the 95\% confidence interval (CI) of precision, recall, f1-score and accuracy as [60.4,61.1], [60.4, 61.1], [61.2, 62.0], and [71.0,71.5], respectively. Also, it has p-value$<$2.2e-16 for precision, recall, f1-score and accuracy results. Based on all of these statistical tests, we believe that our proposed method is robust towards classification.
}

\section{Conclusion and future works}
\label{conclusion_futureworks}
{In this paper, we have proposed a multi-channel convolutional neural network (MCNN) to classify the Nepali COVID-19 related tweets into three different sentiment classes--positive, neutral and negative. Our experimental results on NepCOV19Tweets dataset show that MCNN is able to capture the discriminating information outperforming the state-of-the-art methods. 
To train the proposed MCNN model, we have proposed to use the hybrid feature extraction method (syntactic+semantic), which shows the robust performance against individual methods with the traditional machine learning algorithms.

Our method still has a limited contextual information due to the availability of low resources in Nepali natural language processing. Thus, it would be interesting to combine other kinds of semantic information (e.g., Nepali Wikipedia description) and re-train our model. In addition, the sequence of tokens present in each Nepali tweet could be interesting research to capture the temporal information for their better discrimination.
}

\section{Data availability}
\label{data_availability}
The data used in this study are publicly available at \url{https://www.kaggle.com/mathew11111/nepcov19tweets}.

\section{Conflict of interest}
The authors declare that there is no conflict of interest.
%

\bibliographystyle{spbasic}      
\bibliography{cas-ref}


\end{document}